\documentclass[letterpaper]{article} 
\usepackage{aaai19}  
\usepackage{times}  
\usepackage{helvet}  
\usepackage{courier}  
\usepackage{url}  
\usepackage{graphicx}  

\usepackage{amsmath,amsfonts,amssymb}
\usepackage{paralist}
\usepackage{color,xcolor}
\usepackage{bm}
\usepackage{multirow}
\usepackage{makecell}
\usepackage{caption}
\usepackage[linesnumbered,lined]{algorithm2e}
\usepackage{todonotes}

\newcommand{\rmx}{\mathrm x} 
\newcommand{\tabh}[1]{\multicolumn{1}{c|}{\textbf{#1}}}  
\newcommand{\tabc}[2]{\multicolumn{1}{|c||}{\multirow{#1}{*}{\textbf{#2}}}} 
\newcommand{\loss}[1]{J_{\text{#1}}}
\newcommand{\hyp}[1]{\lambda_{\text{#1}}}
\newcommand{\nnweight}[1]{\bm{\theta_{\text{#1}}}}
\newcommand{\weight}[1]{W_{\text{#1}}}
\newcommand{\bias}[1]{\bm{b_{\text{#1}}}}
\newcommand{\citeay}[1]{\citeauthor{#1} \shortcite{#1}}

\title{Disentangled Representation Learning for Non-Parallel Text Style Transfer}

\author{
	Vineet John \\
	University of Waterloo \\
	{\tt vineet.john@uwaterloo.ca} \\
	\And
	Lili Mou \\
	AdeptMind Research \\
	{\tt doublepower.mou@gmail.com}\\{\tt lili@adeptmind.ai} \\
	\AND
	Hareesh Bahuleyan \\
	University of Waterloo \\
	{\tt hpallika@uwaterloo.ca} \\
	\And
	Olga Vechtomova \\
	University of Waterloo \\
	{\tt ovechtom@uwaterloo.ca} \\
}

\date{}
\begin{document}
\maketitle
\graphicspath{{images/}}

\begin{abstract}
	This paper tackles the problem of disentangling the latent variables of style and content in language models.
	We propose a simple yet effective approach, which incorporates auxiliary multi-task and adversarial objectives, for label prediction and bag-of-words prediction, respectively.
	We show, both qualitatively and quantitatively, that the style and content are indeed disentangled in the latent space.
	This disentangled latent representation learning method is applied to style transfer on non-parallel corpora.
	We achieve substantially better results in terms of transfer accuracy, content preservation and language fluency, in comparison to previous state-of-the-art approaches.\footnote{Our code is publicly available at \url{https://github.com/vineetjohn/linguistic-style-transfer}}
\end{abstract}

\section{Introduction}

The neural network has been a successful learning machine during the past decade due to its highly expressive modeling capability, which is a consequence of multiple layers of non-linear transformations of input features.
Such transformations, however, make intermediate features ``latent,'' in the sense that they do not have explicit meaning and are not interpretable.
Therefore, neural networks are usually treated as black-box machinery.

Disentangling the latent space of neural networks has become an increasingly important research topic.
In the image domain, for example, \citeay{chen2016infogan} use adversarial and information maximization objectives to produce interpretable latent representations that can be tweaked to adjust writing style for handwritten digits, as well as lighting and orientation for face models.
\citeay{mathieu2016disentangling} utilize a convolutional autoencoder to achieve the same objective.
However, this problem is not well explored in natural language processing.

In this paper, we address the problem of disentangling the latent space of neural networks for text generation.
Our model is built on an autoencoder that encodes a sentence to the latent space (vector representation) by learning to reconstruct the sentence itself.
We would like the latent space to be disentangled with respect to different features, namely, \textit{style} and \textit{content} in our task.

To accomplish this, we propose a simple approach that combines multi-task and adversarial objectives.
We artificially divide the latent representation into two parts: the style space and content space. In this work, we consider the sentiment of a sentence as the style.
We design auxiliary losses, enforcing the separation of style and content latent spaces.
In particular, the multi-task loss operates on a latent space to ensure that the space does contain the information we wish to encode.
The adversarial loss, on the contrary, minimizes the predictability of information that should not be contained in that space.
In previous work, researchers typically work with the style, or specifically, sentiment space~\cite{hu2017toward,shen2017style,fu2018style}, but simply ignore the content space, as it is hard to formalize what ``content'' actually refers to.

In our paper, we propose to approximate the content information by bag-of-words (BoW) features, where we focus on style-neutral, non-stopwords.
Along with traditional style-oriented auxiliary losses, our BoW multi-task loss and BoW adversarial loss make the style and content spaces much more disentangled from each other.

The learned disentangled latent space can be directly used for text style-transfer~\cite{hu2017toward,shen2017style}, which aims to transform a given sentence to a new sentence with the same content but a different style.
Since it is difficult to obtain training sentence pairs with the same content and differing styles (i.e. parallel corpora), we follow the setting where we train our model on a non-parallel but style-labeled corpora. We call this \textit{non-parallel text style transfer}.
To accomplish this, we train an autoencoder with disentangled latent spaces.
For style-transfer inference, we simply use the autoencoder to encode the content vector of a sentence, but ignore its encoded style vector.
We then infer from the training data, an empirical embedding of the style that we would like to transfer.
The encoded content vector and the empirically-inferred style vector are concatenated and fed to the decoder.
This grafting technique enables us to obtain a new sentence similar in content to the input sentence, but with a different style.

We conducted experiments on two customer review datasets.
Qualitative and quantitative results show that both the style and content spaces are indeed disentangled well.
In the style-transfer evaluation, we achieve substantially better style-transfer strength, content preservation, and language fluency scores, compared with previous results.
Ablation tests also show that the auxiliary losses can be combined well, each playing its own role in disentangling the latent space.

\section{Related Work}

Disentangling neural networks' latent space has been explored in the image processing domain in the recent years, and researchers have successfully disentangled rotation features, color features, etc.~of images~\cite{chen2016infogan,luan2017deep}.
Some image characteristics (e.g., artistic style) can be captured well by certain statistics \cite{gatys2016image}.
In other work, researchers adopt data augmentation techniques to learn a disentangled latent space~\cite{kulkarni2015deep,champandard2016semantic}.

In natural language processing, the definition of ``style'' itself is vague, and as a convenient starting point, NLP researchers often treat sentiment as a salient style of text.
\citeay{hu2017toward} manage to control the sentiment by using discriminators to reconstruct sentiment and content from generated sentences.
However, there is no evidence that the latent space would be disentangled by this reconstruction.
\citeay{shen2017style} use a pair of adversarial discriminators to align the recurrent hidden decoder states of original and style-transferred sentences, for a given style.
\citeay{fu2018style} propose two approaches: training style-specific embeddings, and training separate style-specific decoders for style-transfer.
They apply an adversarial loss on the encoded space to discourage encoding style in the latent space of an autoencoding model. All the above approaches only deal with the style information and simply ignore the content part.

\citeay{zhao2018adversarially} extend the multi-decoder approach and use a Wasserstein-distance penalty to align content representations of sentences with different styles. However, the Wasserstein penalty is applied to  empirical samples from the data distribution, and is more indirect than our BoW-based auxiliary losses.
Recently, \citeay{rao2018dear} treat the formality of writing as a style, and create a parallel corpus for style transfer with sequence-to-sequence models.
This is beyond the scope of our paper, as we focus on non-parallel text style transfer.

Our paper differs from previous work in that both our style space and content space are encoded from the input, and we design several auxiliary losses to ensure that each space encodes and only encodes the desired information.
Such disentanglement of latent space has its own research interest in the deep learning community.
The disentangled representation can be directly applied to non-parallel text style-transfer tasks, as in the aforementioned studies.

\section{Approach}

\begin{figure}[!t]
	\centering
	\includegraphics[width=.9\linewidth]{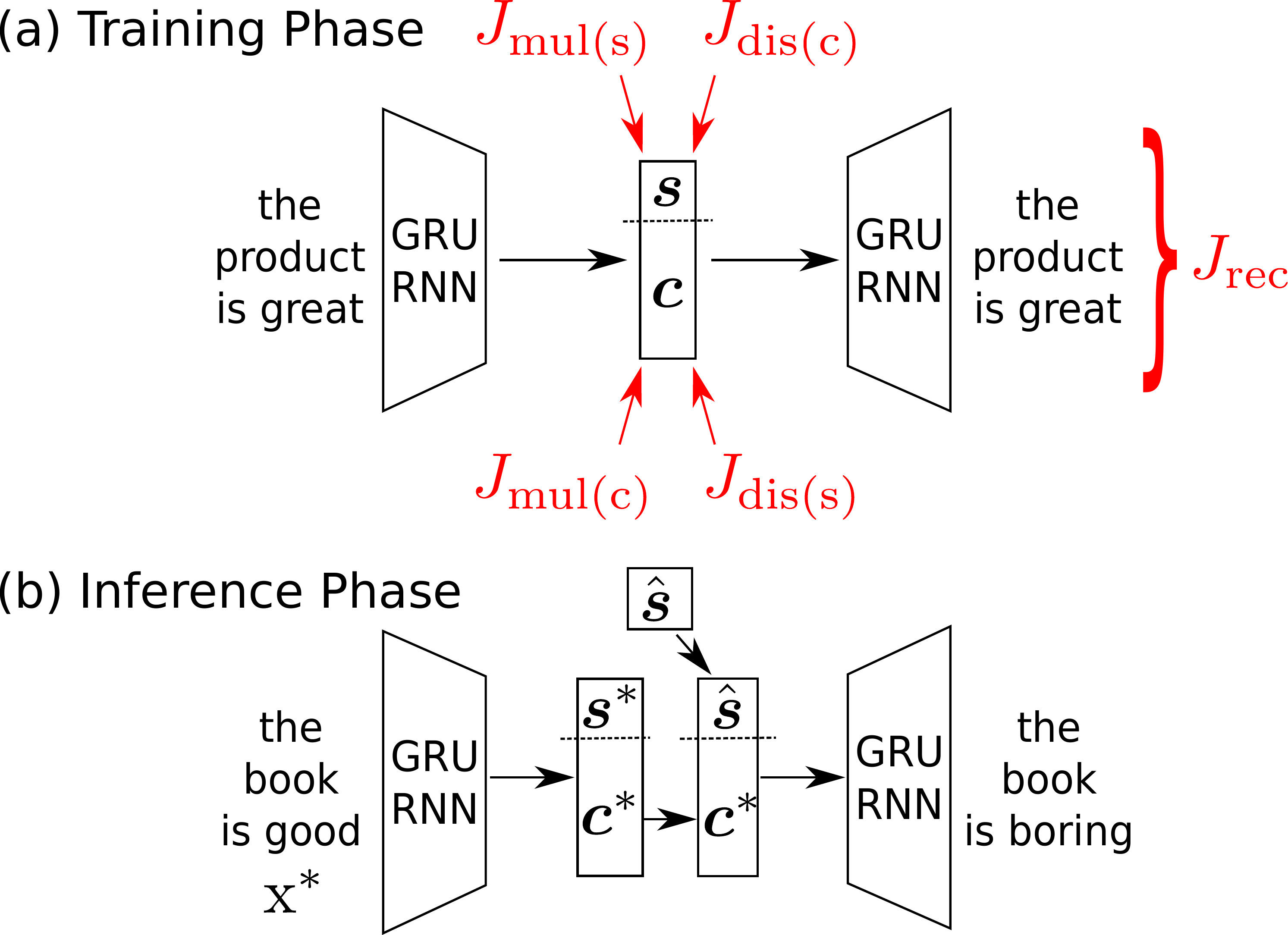}
	\caption{Overview of our approach.}
	\label{fig:overview}
\end{figure}

In this section, we describe our approach in detail, shown in Figure~\ref{fig:overview}.
Our model is built upon an autoencoder with a sequence-to-sequence neural network~\cite{sutskever2014sequence}, and we design multi-task and adversarial losses for both style and content spaces.
Finally, we present our approach to transfer style in the context of natural language generation.

\subsection{Autoencoder} \label{ssec:seq2seq-autoencoder}

An autoencoder encodes an input to a latent vector space, from which it reconstructs the input itself.
The latent vector space is usually of much smaller dimensionality than input data, and the autoencoder learns salient and compact representations of data during the reconstruction process.

Let $\rmx=(x_1, x_2, \cdots x_n)$ be an input sequence with $n$ tokens.
The encoder recurrent neural network (RNN) with gated recurrent units (GRU) \cite{cho2014learning} encodes $\rm x$ and obtains a hidden vector representation $\bm h$, which is linearly transformed from the encoder RNN's final hidden state.

Then a decoder RNN generates a sentence, which ideally should be $\rmx$ itself.
Suppose at a time step $t$, the decoder RNN predicts the word $x_t$ with probability $p(x_t|\bm h, x_1\cdots x_{t-1})$. Then the autoencoder is trained with a sequence-aggregated cross-entropy loss, given by
\begin{equation}\label{eqn:ae}
	\loss{AE}(\nnweight{E},\nnweight{D})= -\sum_{t=1}^n \log p(x_t|\bm h, x_1\cdots x_{t-1})
\end{equation}
where $\nnweight{E}$ and $\nnweight{D}$ are the parameters of the encoder and decoder, respectively.\footnote{For brevity, we only present the loss for a single data point (i.e., a sentence) throughout the paper. The total loss sums over all data points, and is implemented using mini-batches.} Both the encoder and decoder are deterministic functions in the original autoencoder model~\cite{rumelhart1985learning}, and thus we call it a \textit{deterministic autoencoder} (DAE).

\subsubsection{Variational Autoencoder.}

In addition to DAE, we also implement a variational autoencoder (VAE) \cite{kingma2013auto}, which imposes a probabilistic distribution on the latent vector. The Kullback-Leibler (KL) divergence \cite{kullback1951information} penalty is added to the loss function to regularize the latent space. The decoder reconstructs data based on the sampled latent vector from its posterior distribution.

Formally, the autoencoding loss in the VAE is
\begin{align}\label{eqn:vae}
	\loss{AE}(\nnweight{E}, \nnweight{D}) = & - \mathbb{E}_{q_{E}(\bm h|\rmx)} [\log p(\rmx|\bm h)]  \nonumber \\
	                                        & + \hyp{kl}\operatorname{KL}(q_{E}(\bm h|\rmx)\|p(\bm h))
\end{align}
where $\hyp{kl}$ is the hyperparameter balancing the reconstruction loss and the KL term. $p(\bm h)$ is the prior, set to the standard normal distribution $\mathcal{N}(\bm 0,\mathrm I)$. $q_E(\bm h|\mathrm x)$ is the posterior taking the form $\mathcal{N}(\bm \mu,\operatorname{diag} \bm\sigma)$, where $\bm\mu$ and $\bm\sigma$ are predicted by the encoder network.
The motivation for using VAE as opposed to DAE is that the reconstruction is based on the samples of the posterior, which populates encoded vectors to the neighborhood and thus smooths the latent space.
\citeay{bowman2016generating} show that VAE enables more fluent sentence generation from a latent space than DAE.

The autoencoding losses in Equations~(\ref{eqn:ae},\ref{eqn:vae}) serve as our primary training objective.
Besides, the autoencoder is also used for text generation in the style-transfer application.
We also design several auxiliary losses to disentangle the latent space. In particular, we hope that $\bm h$ can be separated into two spaces $\bm s$ and $\bm c$, representing style and content, respectively, i.e., $\bm h = [\bm s ; \bm c]$, where $[\cdot;\cdot]$ denotes concatenation.
This is accomplished by the auxiliary losses described in the rest of this section.

\subsection{Style-Oriented Losses}

We first design auxiliary losses that ensure the style information is contained in the style space $\bm s$.
This involves a multi-task loss that ensures $\bm s$ is discriminative for the style, as well as an adversarial loss that ensures $\bm c$ is not discriminative for the style.

\subsubsection{Multi-Task Loss for Style.} \label{ssec:multitask-style-objective}
Although the corpus we use is non-parallel, we assume that each sentence is labeled with its style. In particular, we treat the sentiment as the style of interest, following previous work~\cite{hu2017toward,shen2017style,fu2018style,zhao2018adversarially}, and each sentence is labeled with a binary sentiment tag (positive or negative).

We build a classifier on the style space that predicts the style label. Formally, a two-way softmax layer (equivalent to logistic regression) is applied to the style vector $\bm s$, given by
\begin{equation} \label{eqn:class-pred}
	\bm y_s = \operatorname{softmax}(\weight{mul(s)} \bm s + \bias{mul(s)})
\end{equation}
where $\nnweight{mul(s)}=[\weight{mul(s)}; \bias{mul(s)}]$ are parameters for multi-task learning of style, and $\bm y_s$ is the output of softmax layer.

The classifier is trained with a simple cross-entropy loss against the ground truth distribution $t_s(\cdot)$, given by
\begin{equation} \label{eqn:style-multi-task-loss}
	\loss{mul(s)}(\nnweight{E};\nnweight{mul(s)}) = - \sum\nolimits_{l\in\text{labels}} t_s(l)\log y_s(l)
\end{equation}
where $\nnweight{E}$ are the encoder's parameters.

We train the style classifier at the same time as the autoencoding loss.
Thus, this could be viewed as \textit{multi-task} learning, incentivizing the entire model to not only decode the sentence, but also predict its sentiment from the style vector $\bm  s$.
We denote it by ``mul(s).''
The idea of multi-task losses is not new and has been used in previous work for sequence-to-sequence learning \cite{luong2015multi}, sentence representation learning \cite{jernite2017discourse} and sentiment analysis \cite{balikas2017multitask}, among others.

\subsubsection{Adversarial Loss for Style.}
\label{ssec:adversarial-style-objective}

The above multi-task loss only ensures that the style space contains style information.
However, the content space might also contain style information, which is undesirable for disentanglement.

We thus apply an adversarial loss to discourage the content space containing style information.
The idea is to first introduce a classifier, called an \textit{adversary}, that deliberately discriminates the true style label using the content vector $\bm c$.
Then the encoder is trained to learn a content vector space, from which its adversary cannot predict style information.

Concretely, the adversarial discriminator and its training objective have a similar form as Equations~\ref{eqn:class-pred} and~\ref{eqn:style-multi-task-loss}, but with different input and parameters, given by
\begin{align}
	\label{eqn:adv-disc-loss}
	\bm y_s                          & = \operatorname{softmax}(\weight{dis(s)} \bm c + \bias{dis(s)}) \\
	\loss{dis(s)}(\nnweight{dis(s)}) & = - \sum\nolimits_{l\in\text{labels}} t_c(l)\log y_s(l)
\end{align}
where $\nnweight{dis(s)}=[\weight{dis(s)}; \bias{dis(s)}]$ are the parameters of the adversary.

It should be emphasized that, for the adversary, the gradients are not propagated back to the autoencoder, i.e., the variables in $\bm c$ are treated as shallow features. Therefore, we view $\loss{dis(s)}$ as a function of $\nnweight{dis(s)}$ only, whereas $\loss{mul(s)}$ is a function of both $\nnweight{E}$ and $\nnweight{mul(s)}$.

Having trained an adversary, we would like the autoencoder to be tuned in such an \textit{ad hoc} fashion, that $\bm c$ is not discriminative for style.
In existing literature, there could be different approaches, for example, maximizing the adversary's loss~\cite{shen2017style,zhao2018adversarially} or penalizing the entropy of the adversary's prediction~\cite{fu2018style}.
In our work, we adopt the latter, as it can be easily extended to multi-category classification, used for the content-oriented losses of our approach. Formally, the adversarial objective for the style is to maximize
\begin{equation} \label{eqn:advs}
	\loss{adv(s)}(\nnweight{E})=\mathcal{H}(\bm y_s|\bm c; \nnweight{dis(s)})
\end{equation}
where $\mathcal{H}(\bm p)=-\sum_{i\in\text{labels}}p_i\log p_i$ and $\bm y_s$ is the predicted distribution over the style labels. Here, $\loss{adv(s)}$ is maximized with respect to the encoder, and attains maximum value when $\bm y_s$ is a uniform distribution. It is viewed as a function of $\nnweight{E}$, and we fix $\nnweight{dis(s)}$.

While adversarial loss has been explored in previous style-transfer papers~\cite{shen2017style,fu2018style}, it has not been combined with the multi-task loss. As we shall show in our experiments, combining these two losses is promisingly effective, achieving better style transfer performance than a variety of previous state-of-the-art methods.

\subsection{Content-Oriented Losses}

The above style-oriented losses only regularize style information, but they do not impose any constraint on how the content information should be encoded. This also happens in most previous work~\cite{hu2017toward,shen2017style,fu2018style}.
Although the style space is usually much smaller than the content space, it is unrealistic to expect that the content would not flow into the style space because of its limited capacity. Therefore, we need to design content-oriented auxiliary losses to regularize the content information.

Inspired by the above combination of multi-task and adversarial losses, we apply the same idea to the content space. However, it is usually hard to define what ``content'' actually refers to.

To this end, we propose to approximate the content information by bag-of-words (BoW) features.
The BoW features of an input sentence is a vector, each element indicating the probability of a word's occurrence in the sentence.
For a sentence $\rmx$ with $N$ words, the word $w_*$'s BoW probability is given by
$t_c(w_*)=\frac{\sum_{i=1}^{N}{\mathbb{I}\{w_i = w_*\}}}{N}$,
where $t_c(\cdot)$ denotes the target distribution of content, and $\mathbb{I\{\cdot\}}$ is an indicator function.
Here, we only consider content words, excluding stopwords and style-specific words, since we focus on ``content'' information. In particular, we exclude sentiment words from a curated lexicon \cite{hu2004mining} for sentiment style transfer.
The effect of using different vocabularies for BoW is analyzed in Supplemental Material A.

\subsubsection{Multi-Task Loss for Content.} \label{ssec:multitask-content-objective}

Similar to the style-oriented loss, the multi-task loss for content, denoted as ``mul(c)'', ensures that the content space $\bm c$ contains content information, i.e., BoW features.

We introduce a softmax classifier over the BoW vocabulary
\begin{equation} \label{eqn:bow-pred}
	\bm y_c = \operatorname{softmax}({\weight{mul(c)}} \bm c + \bias{mul(c)})
\end{equation}
where $\nnweight{mul(c)}=[\weight{mul(c)}; \bias{mul(c)}]$ are the classifier's parameters, and $\bm y_c$ is the predicted BoW distribution.

The training objective is a cross-entropy loss against the ground truth distribution $t_c(\cdot)$, given by
\begin{equation}\label{eqn:content-multi-task-loss}
	\loss{mul(c)}(\nnweight{E};\nnweight{mul(c)}) = - \sum\nolimits_{w\in\text{vocab}} t_c(w)\log y_c(w)
\end{equation}
where the optimization is performed with both encoder parameters $\nnweight{E}$ and the multi-task classifier $\nnweight{mul(c)}$.
Notice that although the target distribution is not one-hot as for BoW prediction, the cross-entropy loss (Equation~\ref{eqn:content-multi-task-loss}) has the same form.

It is also interesting that, at first glance, the multi-task loss for content appears to be redundant, given the autoencoding loss, when in fact, it is not. The multi-task loss only considers content words, which do not include stopwords and sentiment words, and is only applied to the content space $\bm c$. This ensures that the content information is captured in the content space. The autoencoding loss only requires that the model reconstructs the sentence based on the content and style space, and does not ensure their separation.

\subsubsection{Adversarial Loss for Content.} \label{ssec:adversarial-content-objective}

To ensure that the style space does not contain content information, we design our final auxiliary loss, the adversarial loss for content, denoted as ``adv(c).''

We build an adversary, a softmax classifier on the style space to predict BoW features, approximating content information, given by
\begin{align}
	\label{eqn:adv-bow-disc-loss}
	\bm y_c                          & = \operatorname{softmax}({\weight{dis(c)}}^\top \bm s + \bias{dis(c)}) \\
	\loss{dis(c)}(\nnweight{dis(c)}) & = - \sum\nolimits_{w\in\text{vocab}} t_c(w)\log y_c(w)
\end{align}
where $\nnweight{dis(c)}=[\weight{dis(c)}; \bias{dis(c)}]$ are the classifier's parameters for BoW prediction.

The adversarial loss for the model is to maximize the entropy of the discriminator
\begin{equation}
	\loss{adv(c)}(\nnweight{E}) = \mathcal{H}(\bm y_c | \bm s; \nnweight{dis(c)})
\end{equation}
Again, $\loss{dis(c)}$ is trained with respect to the discriminator's parameters $\nnweight{dis(c)}$, whereas $\loss{adv(c)}$ is trained with respect to $\nnweight{E}$, similar to the adversarial loss for style.

\subsection{Training Process}

The overall loss $\loss{ovr}$ for the autoencoder comprises several terms: the reconstruction objective, the multi-task objectives for style and content, and the adversarial objectives for style and content:
\begin{align}
	\loss{ovr} = & \loss{AE}(\nnweight{E}, \nnweight{D})                                                                              \\
	             & + \hyp{mul(s)} \loss{mul(s)} (\nnweight{E},\nnweight{mul(s)}) - \hyp{adv(s)} \loss{adv(s)}(\nnweight{E}) \nonumber \\
	             & + \hyp{mul(c)} \loss{mul(c)} (\nnweight{E},\nnweight{mul(c)}) - \hyp{adv(c)} \loss{adv(c)}(\nnweight{E}) \nonumber
\end{align}
where $\lambda$'s are the hyperparameters that balance the autoencoding loss and these auxiliary losses.

To put it all together, the model training involves an alternation of optimizing discriminator losses $\loss{dis(s)}$ and $\loss{dis(c)}$, and the model's own loss $\loss{ovr}$, shown in Algorithm~\ref{alg:training-process}.

\begin{algorithm}[!t]
	\ForEach{mini-batch}{
		minimize $\loss{dis(s)}(\nnweight{dis(s)})$ w.r.t. $\nnweight{dis(s)}$\;
		minimize $\loss{dis(c)}(\nnweight{dis(c)})$ w.r.t. $\nnweight{dis(c)}$\;
		minimize $\loss{ovr}$ w.r.t. $\nnweight{E}, \nnweight{D}, \nnweight{mul(s)}, \nnweight{mul(c)}$\;
	}
	\caption{Training process.}\vspace{-.2cm}
	\label{alg:training-process}
\end{algorithm}

\subsection{Generating Style-Transferred Sentences} \label{ssec:sentence-generation}

A direct application of our disentangled latent space is style-transfer for natural language generation.
For example, we can generate a sentence with generally the same meaning (content) but a different style (e.g., sentiment).

Let $\rmx^*$ be an input sentence with $\bm s^*$ and $\bm c^*$ being the encoded, disentangled style and content vectors, respectively.
If we would like to transfer its content to a different style, we compute an empirical estimate of the target style's vector $\hat{\bm s}$ using
\begin{equation*}
	\hat{\bm s}=\frac{\sum_{i\in\text{target style}}\bm s_i}{\text{\# target style samples}}
\end{equation*}
The inferred target style $\hat{\bm s}$ is concatenated with the encoded content $\bm c^*$ for decoding style-transferred sentences, as shown in Figure~\ref{fig:overview}b.

\section{Experiments}

\subsection{Datasets}

We conducted experiments on two datasets, Yelp and Amazon reviews.
Both of these datasets comprise sentences accompanied by binary sentiment labels (positive, negative). They are used to train latent space disentanglement as well as to evaluate sentiment transfer.

\subsubsection{Yelp Service Reviews.}
We used a Yelp review dataset, following previous work \cite{shen2017style,zhao2018adversarially}.\footnote{The Yelp dataset is available at \url{https://github.com/shentianxiao/language-style-transfer}}
It contains 444101, 63483 and 126670 labeled reviews for train, validation, and test, respectively.
The maximum review length is 15 words, and the vocabulary size is approximately 9200.

\subsubsection{Amazon Product Reviews.}
We further evaluate our model with an Amazon review dataset, following another previous paper~\cite{fu2018style}.\footnote{The Amazon dataset is available at \url{https://github.com/fuzhenxin/text_style_transfer}}
It contains 559142, 2000 and 2000 labeled reviews for train, validation, and test, respectively.
The maximum review length is 20 words, and the vocabulary size is approximately 58000.

\subsection{Experiment Settings}
We used the Adam optimizer \cite{kingma2014adam} for the autoencoder and the RMSProp optimizer \cite{tieleman2012lecture} for the discriminators, following adversarial training stability tricks \cite{arjovsky2017wasserstein}.
Each optimizer has an initial learning rate of $10^{-3}$.
Our model is trained for 20 epochs, by which time it has mostly converged.
The word embedding layer was initialized by word2vec \cite{mikolov2013distributed} trained on respective training sets.
Both the autoencoder and the discriminators are trained once per mini-batch with $\hyp{mul(s)} = 10$, $\hyp{mul(c)} = 3$, $\hyp{adv(s)} = 1$, and $\hyp{adv(c)} = 0.03$.
These hyperparameters were tuned by performing a log-scale grid search within two orders of magnitude around the default value $1$, and choosing those that yielded the best validation results.
The recurrent unit size is $256$, the style vector size is $8$, and the content vector size is $128$.
We append the latent vector $\bm h$ to the hidden state at every time step of the decoder.

For the VAE model, we enforce the KL-divergence penalty on both the style and content posterior distributions, using $\hyp{kl(s)}$ and $\hyp{kl(c)}$, respectively.
We set $\hyp{kl(s)} = 0.03$ and $\hyp{kl(c)} = 0.03$ and use the $\operatorname{sigmoid}$ KL-weight annealing schedule following \citeay{bahuleyan2018probabilistic}.
They were tuned in the same manner as the other hyperparameters of the model.

\subsection{Experiment I: Disentangling Latent Space}

First, we analyze how the style (sentiment) and content of the latent space are disentangled.
We train classifiers on the different latent spaces, and report their inference-time classification accuracies in Table~\ref{tab:classification-yelp}.

We see that the 128-dimensional content vector $\bm c$ is not particularly discriminative for style.
It achieves accuracies slightly better than majority guess.
However, the 8-dimensional style vector $\bm s$, despite its low dimensionality, achieves substantially higher style classification accuracy.
When combining content and style vectors, we observe no further improvement.
These results verify the effectiveness of our disentangling approach, as the style space contains style information, whereas the content space does not.

\begin{figure}[!t]
	\centering
	\includegraphics[width=\linewidth]{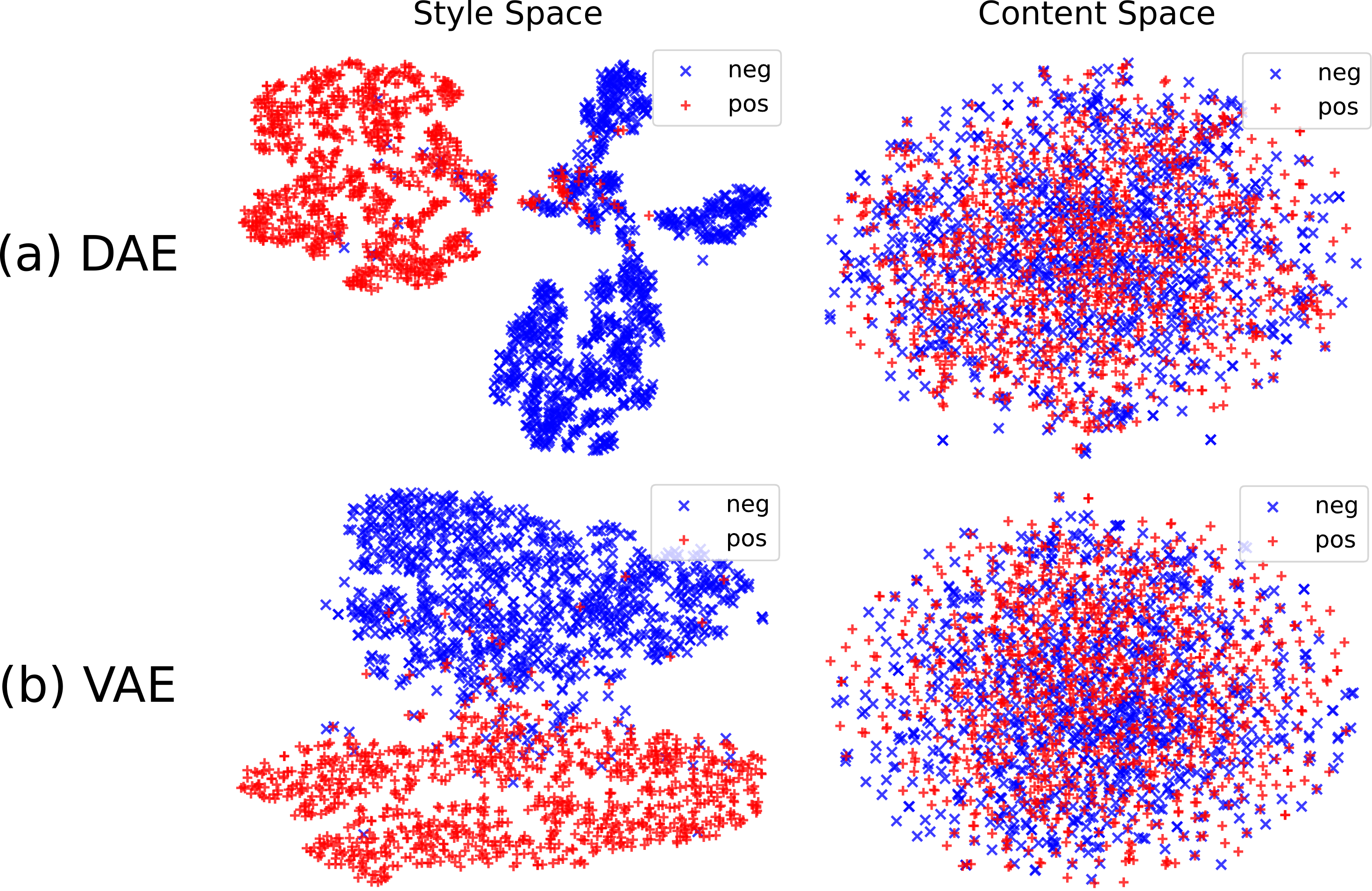}
	\caption{t-SNE plots of the disentangled style and content spaces (with all auxiliary losses on the Yelp dataset).}
	\label{fig:tsne-plots}
\end{figure}

We show t-SNE plots of both the deterministic autoencoder (DAE) and the variational autoencoder (VAE) models in Figure~\ref{fig:tsne-plots}.
As seen, sentences with different styles are noticeably separated in a clean manner in the style space (LHS), but are indistinguishable in the content space (RHS).
It is also evident that the latent space learned by the variational autoencoder is considerably smoother and continuous compared with the one learned by the deterministic autoencoder.

We show t-SNE plots for ablation tests with different combinations of auxiliary losses in Supplemental Material B.

\begin{table}[!t]
	\centering
	\resizebox{\linewidth}{!}{
		\begin{tabular}{|l||r|r||r|r|}
			\hline
			\multirow{2}{*}{\textbf{Latent Space}} & \multicolumn{2}{c||}{\textbf{Yelp}} & \multicolumn{2}{c|}{\textbf{Amazon}}                             \\
			\cline{2-5}
			                                       & \textbf{DAE}                        & \textbf{VAE}                         & \textbf{DAE} & \tabh{VAE} \\
			\hline \hline
			None (majority guess)                  & \multicolumn{2}{c||}{0.602}         & \multicolumn{2}{c|}{0.512}                                       \\
			\hline
			Content space  ($\bm c$)               & 0.658                               & 0.697                                & 0.675        & 0.693      \\ \hline
			Style space ($\bm s$)                  & 0.974                               & 0.974                                & 0.821        & 0.810      \\ \hline
			Complete space ($[\bm s;\bm c]$)       & 0.974                               & 0.974                                & 0.819        & 0.810      \\
			\hline
		\end{tabular}}
	\caption{Classification accuracy on latent spaces.}
	\label{tab:classification-yelp}
\end{table}

\begin{table*}[!t]
	\centering
	\resizebox{\textwidth}{!}{
		\begin{tabular}{|l||c|c|c|c||c|c|c|c| }
			\hline
			\tabc{3}{Model}                            & \multicolumn{4}{c||}{\textbf{Yelp Dataset}} & \multicolumn{4}{c|}{\textbf{Amazon Dataset}}                                                                                                                             \\
			\cline{2-9}
			                                           & \textbf{Transfer}                           & \textbf{Cosine}                              & \textbf{Word}    & \textbf{Language} & \textbf{Transfer} & \textbf{Cosine}         & \textbf{Word}    & \textbf{Language} \\
			                                           & \textbf{Accuracy}                           & \textbf{Similarity}                          & \textbf{Overlap} & \textbf{Fluency}  & \textbf{Accuracy} & \textbf{Similarity}     & \textbf{Overlap} & \textbf{Fluency}  \\
			\hline
			\hline
			Style-Embedding \cite{fu2018style}         & 0.182                                       & \textbf{0.959}                               & \textbf{0.666}   & -16.17            & \ 0.400$^\dag$    & \ \textbf{0.930}$^\dag$ & \textbf{0.359}   & -28.13            \\
			\hline		\hline
			Cross-Alignment \cite{shen2017style}       & \ 0.784$^\dag$                              & 0.892                                        & 0.209            & -23.39            & 0.606             & 0.893                   & 0.024            & -26.31            \\
			\hline
			Multi-Decoder \cite{zhao2018adversarially} & \ 0.818$^\dag$                              & 0.883                                        & 0.272            & -20.95            & 0.552             & \textit{0.926}          & 0.169            & -34.70            \\
			\hline
			Ours (DAE)                                 & 0.883                                       & \textit{0.915}                               & \textit{0.549}   & -10.17            & 0.720             & 0.921                   & \textit{0.354}   & -24.74            \\
			\hline
			Ours (VAE)                                 & \textbf{0.934}                              & {0.904}                                      & {0.473}          & \textbf{-9.84}    & \textbf{0.822}    & 0.900                   & 0.196            & \textbf{-21.70}   \\
			\hline
		\end{tabular}}\vspace{-.2cm}
	\caption{Performance of non-parallel text style transfer. The style-embedding approach achieves poor transfer accuracy, and should not be considered as an effective style-transfer model. Despite this, our model outperforms other previous methods in terms of all aspects (transfer strength, content preservation, and language fluency). Numbers with the $^\dag$ symbol are quoted from respective papers. Others are based on our replication using the published code in previous work. Our replicated experiments achieve 0.809 and 0.835 transfer accuracy on the Yelp dataset, close to the results in \protect\citeay{shen2017style} and \citeay{zhao2018adversarially}, respectively, showing that our replication is fair.}\vspace{-.2cm}
	\label{tab:yelp-comparison-previous}
\end{table*}

\subsection{Experiment II: Non-Parallel Text Style Transfer}
We also conducted sentiment transfer experiments with our disentangled latent space.

\subsubsection{Metrics.} We evaluate competing models based on (1) style transfer strength, (2) content preservation and (3) quality of generated language. The evaluation of generated sentences is a difficult task in contemporary literature, so we adopt a few automatic metrics and use human judgment as well.

$\bullet$ \textit{Style-Transfer Accuracy.} We follow most previous work~\cite{hu2017toward,shen2017style,fu2018style} and train a separate convolutional neural network (CNN)  to predict the sentiment of a sentence~\cite{kim2014convolutional}, which is then used to approximate the style transfer accuracy. In other words, we report the CNN classifier's accuracy on the style-transferred sentences, considering the target style to be the ground truth.

While the style classifier itself may not be perfect, it achieves a reasonable sentiment accuracy on the validation sets ($97\%$ for Yelp; $82\%$ for Amazon). Thus, it provides a quantitative way of evaluating the strength of style-transfer.

$\bullet$ \textit{Cosine Similarity.}
We followed \citeay{fu2018style} and computed a sentence embedding by concatenating the $\operatorname{min}$, $\operatorname{max}$, and $\operatorname{mean}$ of its word embeddings (sentiment words removed).
Then, we computed the cosine similarity between the source and generated sentence embeddings, which is intended to be an indicator of content preservation.

$\bullet$ \textit{Word Overlap.} We find that the cosine similarity measure, although correlated to human judgment, is not a sensitive measure, and we propose a simple yet effective measure that counts the unigram word overlap rate of the original sentence $\mathrm x$ and the style-transferred sentence $\mathrm y$, computed by $\frac{count(w_{\mathrm x} \cap w_{\mathrm y})}{count(w_{\mathrm x} \cup w_{\mathrm y})}$.

$\bullet$ \textit{Language Fluency.}
We use a trigram Kneser-Ney (KL) smoothed language model \cite{kneser1995improved} as a quantitative and automated metric to evaluate the fluency of a sentence.
It estimates the empirical distribution of trigrams in a corpus, and computes the log-likelihood of a test sentence.
We train the language model on the respective dataset, and report the Kneser-Ney language model's log-likelihood. A larger (closer to zero) number indicates a more fluent sentence.

$\bullet$  \textit{Manual Evaluation.}
Despite the above automatic metrics, we also conduct human evaluations to further confirm the performance of our model.
This was done on the Yelp dataset only, due to the amount of manual effort involved.
We asked 6 human evaluators to rate each sentence on a 1--5 Likert scale~\cite{stent2005evaluating} in terms of transfer strength, content similarity, and language quality.
This evaluation was conducted in a strictly blind fashion: samples obtained from all evaluated models are randomly shuffled, so that the evaluator would be unaware of which model generated a particular sentence.
The inter-rater agreement---as measured by Krippendorff's alpha \cite{krippendorf2004content} for our Likert scale ratings---is 0.74, 0.68, and 0.72 for transfer strength, content preservation, and language quality, respectively.
According to \citeay{krippendorf2004content}, this is an acceptable inter-rater agreement.
\begin{table}[!t]
	\centering\vspace{-.2cm}
	\resizebox{\linewidth}{!}{
		\begin{tabular}{| l || c | c | c | }
			\hline
			\tabc{2}{Model}                & \tabh{Transfer} & \tabh{Content}      & \tabh{Language} \\
			                               & \tabh{Strength} & \tabh{Preservation} & \tabh{Quality}  \\
			\hline
			\hline		\citeay{fu2018style}     & 1.67            & \textbf{3.84}       & 3.66            \\\hline\hline
			\citeay{shen2017style}         & 3.63            & 3.07                & 3.08            \\
			\hline
			\citeay{zhao2018adversarially} & 3.55            & 3.09                & 3.77            \\
			\hline
			Ours (DAE)                     & 3.67            & 3.64                & 4.19            \\
			\hline
			Ours (VAE)                     & \textbf{4.32}   & \textit{3.73}       & \textbf{4.48}   \\
			\hline
		\end{tabular}}
	\caption{Manual evaluation on the Yelp dataset.}
	\label{tab:manual-evaluation}
\end{table}

\subsubsection{Results and Analysis.}

We compare our approach with previous state-of-the-art work in Table \ref{tab:yelp-comparison-previous}.
For baseline methods, we quoted results from existing papers whenever possible, and replicated the experiments to report other metrics with publicly available code \cite{shen2017style,fu2018style,zhao2018adversarially}.\footnote{\citeay{fu2018style} propose another model using multiple decoders; the method is further developed in \citeay{zhao2018adversarially}, and we adopt the latter for comparison.} As discussed in Table~\ref{tab:yelp-comparison-previous}, our replication involves reasonable efforts and is fair for comparison.

\begin{table*}[ht]
	\centering\vspace{-.2cm}
	\begin{tabular}{| l || c | c | c | c |}
		\hline
		\tabc{2}{Objectives}                                                             & \tabh{Transfer} & \tabh{Cosine}     & \tabh{Word}    & \tabh{Language} \\
		                                                                                 & \tabh{Accuracy} & \tabh{Similarity} & \tabh{Overlap} & \tabh{Fluency}  \\
		\hline
		\hline
		$\loss{AE}$                                                                      & 0.106           & \textbf{0.939}    & 0.472          & -12.58          \\
		\hline
		$\loss{AE}$, $\loss{mul(s)}$                                                     & 0.767           & 0.911             & 0.331          & -12.17          \\
		\hline
		$\loss{VAE}$, $\loss{adv(s)}$                                                    & 0.782           & 0.886             & 0.230          & -12.03          \\
		\hline
		$\loss{VAE}$, $\loss{mul(s)}$, $\loss{adv(s)}$                                   & 0.912           & 0.866             & 0.171          & -9.59           \\
		\hline
		$\loss{VAE}$, $\loss{mul(s)}$, $\loss{adv(s)}$, $\loss{mul(c)}$, $\loss{adv(c)}$ & \textbf{0.934}  & 0.904             & \textbf{0.473} & \textbf{-9.84}  \\
		\hline
	\end{tabular}\vspace{-.2cm}
	\caption{Ablation tests on the Yelp dataset. In all variants, we follow the same protocol of style transfer by substituting an empirical estimate of the target style vector.}
	\label{tab:ablation-results}
\end{table*}
\begin{table*}[!t]
	\centering
	\resizebox{.95\textwidth}{!}{\footnotesize
		\begin{tabular}{| p{0.3\linewidth} || p{0.3\linewidth} | p{0.3\linewidth} |}
			\hline
			\tabc{1}{Original (Positive)}                                           & \tabh{DAE Transferred (Negative)}                                         & \tabh{VAE Transferred (Negative)}                          \\
			\hline
			\hline
			the food is excellent and the service is exceptional                    & the food was a bit bad but the staff was exceptional                      & the food was bland and i am not thrilled with this         \\
			\hline
			the waitresses are friendly and helpful                                 & the guys are rude and helpful                                             & the waitresses are rude and are lazy                       \\
			\hline
			the restaurant itself is romantic and quiet                             & the restaurant itself is awkward and quite crowded                        & the restaurant itself was dirty                            \\
			\hline
			great deal                                                              & horrible deal                                                             & no deal                                                    \\
			\hline
			both times i have eaten the lunch buffet and it was outstanding         & their burgers were decent but the eggs were not the consistency           & both times i have eaten here the food was mediocre at best \\
			\hline
			\hline
			\tabc{1}{Original (Negative)}                                           & \tabh{DAE Transferred (Positive)}                                         & \tabh{VAE Transferred (Positive)}                          \\
			\hline
			\hline
			the desserts were very bland                                            & the desserts were very good                                               & the desserts were very good                                \\
			\hline
			it was a bed of lettuce and spinach with some italian meats and cheeses & it was a beautiful setting and just had a large variety of german flavors & it was a huge assortment of flavors and italian food       \\
			\hline
			the people behind the counter were not friendly whatsoever              & the best selection behind the register and service presentation           & the people behind the counter is friendly caring           \\
			\hline
			the interior is old and generally falling apart                         & the decor is old and now perfectly                                        & the interior is old and noble                              \\
			\hline
			they are clueless                                                       & they are stoked                                                           & they are genuinely professionals                           \\
			\hline
		\end{tabular}}\vspace{-.2cm}
	\caption{Examples of style transferred sentence generation.}\vspace{-.2cm}
	\label{tab:transfer-samples}
\end{table*}

We observe that the style embedding model \cite{fu2018style} performs poorly on the style-transfer objective,\footnote{It should be noted that the transfer accuracy is lower bounded by 0\% as opposed to 50\%, because we always transfer a sentence to the opposite sentiment. The lower-bound, zero transfer accuracy, is achieved by a trivial model that copies the input.} resulting in inflated cosine similarity and word overlap scores. We also examined the number of times each model generates exact copies of the source sentences during style transfer. We notice that the style-embedding model simply reconstructs the exact source sentence $24\%$ of the time, whereas all other models do this less than $6\%$ of the time. Therefore, we do not think that the style embedding approach is an effective model for text style transfer.

The other two competing methods~\cite{shen2017style,zhao2018adversarially} achieve reasonable transfer accuracy and cosine similarity. However, our model outperforms them by 10\% transfer accuracy as well as content preserving scores (measured by cosine similarity and the word overlap rate). This shows our model is able to generate high-quality style transferred sentences, which in turn indicates that the  latent space  is well disentangled into style and content subspaces. Regarding language fluency, we see that VAE is better than DAE in both experiments. This is expected as VAE regularizes the latent space by imposing a probabilistic distribution. We also see that our method achieves considerably more fluent sentences than competing methods, showing that our multi-task and adversarial losses are more ``natural'' than other methods, for example, aligning RNN hidden states \cite{shen2017style}.

Table~\ref{tab:manual-evaluation} presents the results of human evaluation. Again, we see that the style embedding model \cite{fu2018style} is ineffective as it has a very low transfer strength, and that our method outperforms other baselines in all aspects. The results are consistent with the automatic metrics in both experiments (Table~\ref{tab:yelp-comparison-previous}). This implies that the automatic metrics we used are reasonable; it also shows consistent evidence of the effectiveness of our approach.

We conducted ablation tests on the Yelp dataset, and show results in Table~\ref{tab:ablation-results}. With $\loss{VAE}$ only, we cannot achieve reasonable style transfer accuracy by substituting an empirically estimated style vector of the target style.  This is because the style and content spaces would not be disentangled spontaneously with the autoencoding loss alone.

With either $\loss{mul(s)}$ or $\loss{adv(s)}$, the model achieves reasonable transfer accuracy and cosine similarity. Combining them together improves the transfer accuracy to 90\%, outperforming previous methods by a margin of 10\% (Table~\ref{tab:yelp-comparison-previous}). This shows that the multi-task loss and the adversarial loss work in different ways. Our insight of combining the two auxiliary losses is a simple yet effective way of disentangling latent space.

However, $\loss{mul(s)}$ and $\loss{adv(s)}$ only regularize the style information, leading to gradual drop of content preserving scores. Then, we have another insight of introducing content-oriented auxiliary losses, $\loss{mul(c)}$ and $\loss{adv(c)}$, based on BoW features, which regularize the content information in the same way as the style information. By incorporating all these auxiliary losses, we achieve high transfer accuracy, high content preservation, as well as high language fluency.

Table~\ref{tab:transfer-samples} provides several examples of our style-transfer model. Results show that we can successfully transfer the sentiment while preserving the content of a sentence.
We see that, with the empirically estimated style vector, we can reliably control the sentiment of generated sentences.

\section{Conclusion}
In this paper, we propose a simple yet effective approach for disentangling the latent space of neural networks.
We combine multi-task and adversarial objectives to separate content and style information from each other, and propose to approximate content information with bag-of-words features of style-neutral, non-stopword vocabulary.

Both qualitative and quantitative experiments show that the latent space is indeed separated into style and content parts.
This disentangled space can be directly applied to text style-transfer tasks.
It achieves substantially better style-transfer strength, content-preservation scores, as well as language fluency, compared with previous state-of-the-art work.

\bibliography{main}

\begin{thebibliography}{}

\bibitem[\protect\citeauthoryear{Arjovsky, Chintala, and
  Bottou}{2017}]{arjovsky2017wasserstein}
Arjovsky, M.; Chintala, S.; and Bottou, L.
\newblock 2017.
\newblock Wasserstein generative adversarial networks.
\newblock In {\em ICML},  214--223.

\bibitem[\protect\citeauthoryear{Bahuleyan \bgroup et al\mbox.\egroup
  }{2018}]{bahuleyan2018probabilistic}
Bahuleyan, H.; Mou, L.; Vamaraju, K.; Zhou, H.; and Vechtomova, O.
\newblock 2018.
\newblock Probabilistic natural language generation with wasserstein
  autoencoders.
\newblock {\em arXiv preprint arXiv:1806.08462}.

\bibitem[\protect\citeauthoryear{Balikas, Moura, and
  Amini}{2017}]{balikas2017multitask}
Balikas, G.; Moura, S.; and Amini, M.-R.
\newblock 2017.
\newblock Multitask learning for fine-grained twitter sentiment analysis.
\newblock In {\em SIGIR},  1005--1008.

\bibitem[\protect\citeauthoryear{Bowman \bgroup et al\mbox.\egroup
  }{2016}]{bowman2016generating}
Bowman, S.~R.; Vilnis, L.; Vinyals, O.; Dai, A.; Jozefowicz, R.; and Bengio, S.
\newblock 2016.
\newblock Generating sentences from a continuous space.
\newblock In {\em CoNLL},  10--21.

\bibitem[\protect\citeauthoryear{Champandard}{2016}]{champandard2016semantic}
Champandard, A.~J.
\newblock 2016.
\newblock Semantic style transfer and turning two-bit doodles into fine
  artworks.
\newblock {\em arXiv preprint arXiv:1603.01768}.

\bibitem[\protect\citeauthoryear{Chen \bgroup et al\mbox.\egroup
  }{2016}]{chen2016infogan}
Chen, X.; Duan, Y.; Houthooft, R.; Schulman, J.; Sutskever, I.; and Abbeel, P.
\newblock 2016.
\newblock Infogan: Interpretable representation learning by information
  maximizing generative adversarial nets.
\newblock In {\em NIPS},  2172--2180.

\bibitem[\protect\citeauthoryear{Cho \bgroup et al\mbox.\egroup
  }{2014}]{cho2014learning}
Cho, K.; van Merrienboer, B.; Gulcehre, C.; Bahdanau, D.; Bougares, F.;
  Schwenk, H.; and Bengio, Y.
\newblock 2014.
\newblock Learning phrase representations using rnn encoder--decoder for
  statistical machine translation.
\newblock In {\em EMNLP},  1724--1734.

\bibitem[\protect\citeauthoryear{Fu \bgroup et al\mbox.\egroup
  }{2018}]{fu2018style}
Fu, Z.; Tan, X.; Peng, N.; Zhao, D.; and Yan, R.
\newblock 2018.
\newblock Style transfer in text: Exploration and evaluation.
\newblock In {\em AAAI},  663--670.

\bibitem[\protect\citeauthoryear{Gatys, Ecker, and
  Bethge}{2016}]{gatys2016image}
Gatys, L.~A.; Ecker, A.~S.; and Bethge, M.
\newblock 2016.
\newblock Image style transfer using convolutional neural networks.
\newblock In {\em CVPR},  2414--2423.

\bibitem[\protect\citeauthoryear{Hu and Liu}{2004}]{hu2004mining}
Hu, M., and Liu, B.
\newblock 2004.
\newblock Mining and summarizing customer reviews.
\newblock In {\em KDD},  168--177.

\bibitem[\protect\citeauthoryear{Hu \bgroup et al\mbox.\egroup
  }{2017}]{hu2017toward}
Hu, Z.; Yang, Z.; Liang, X.; Salakhutdinov, R.; and Xing, E.~P.
\newblock 2017.
\newblock Toward controlled generation of text.
\newblock In {\em ICML},  1587--1596.

\bibitem[\protect\citeauthoryear{Jernite, Bowman, and
  Sontag}{2017}]{jernite2017discourse}
Jernite, Y.; Bowman, S.~R.; and Sontag, D.
\newblock 2017.
\newblock Discourse-based objectives for fast unsupervised sentence
  representation learning.
\newblock {\em arXiv preprint arXiv:1705.00557}.

\bibitem[\protect\citeauthoryear{Kim}{2014}]{kim2014convolutional}
Kim, Y.
\newblock 2014.
\newblock Convolutional neural networks for sentence classification.
\newblock In {\em EMNLP},  1746--1751.

\bibitem[\protect\citeauthoryear{Kingma and Ba}{2014}]{kingma2014adam}
Kingma, D.~P., and Ba, J.
\newblock 2014.
\newblock Adam: A method for stochastic optimization.
\newblock {\em arXiv preprint arXiv:1412.6980}.

\bibitem[\protect\citeauthoryear{Kingma and Welling}{2014}]{kingma2013auto}
Kingma, D.~P., and Welling, M.
\newblock 2014.
\newblock Auto-encoding variational bayes.
\newblock {\em International Conference on Learning Representations}.

\bibitem[\protect\citeauthoryear{Kneser and Ney}{1995}]{kneser1995improved}
Kneser, R., and Ney, H.
\newblock 1995.
\newblock Improved backing-off for m-gram language modeling.
\newblock In {\em icassp}, volume~1,  181e4.

\bibitem[\protect\citeauthoryear{Krippendorf}{2004}]{krippendorf2004content}
Krippendorf, K.
\newblock 2004.
\newblock {\em Content analysis: An introduction to its methodology}.
\newblock London: SAGE.

\bibitem[\protect\citeauthoryear{Kulkarni \bgroup et al\mbox.\egroup
  }{2015}]{kulkarni2015deep}
Kulkarni, T.~D.; Whitney, W.~F.; Kohli, P.; and Tenenbaum, J.
\newblock 2015.
\newblock Deep convolutional inverse graphics network.
\newblock In {\em NIPS},  2539--2547.

\bibitem[\protect\citeauthoryear{Kullback and
  Leibler}{1951}]{kullback1951information}
Kullback, S., and Leibler, R.~A.
\newblock 1951.
\newblock On information and sufficiency.
\newblock {\em The Annals of Mathematical Statistics} 22(1):79--86.

\bibitem[\protect\citeauthoryear{Luan \bgroup et al\mbox.\egroup
  }{2017}]{luan2017deep}
Luan, F.; Paris, S.; Shechtman, E.; and Bala, K.
\newblock 2017.
\newblock Deep photo style transfer.
\newblock In {\em Proceedings of the IEEE Conference on Computer Vision and
  Pattern Recognition},  4990--4998.

\bibitem[\protect\citeauthoryear{Luong \bgroup et al\mbox.\egroup
  }{2015}]{luong2015multi}
Luong, M.-T.; Le, Q.~V.; Sutskever, I.; Vinyals, O.; and Kaiser, L.
\newblock 2015.
\newblock Multi-task sequence to sequence learning.
\newblock {\em arXiv preprint arXiv:1511.06114}.

\bibitem[\protect\citeauthoryear{Mathieu \bgroup et al\mbox.\egroup
  }{2016}]{mathieu2016disentangling}
Mathieu, M.~F.; Zhao, J.~J.; Zhao, J.; Ramesh, A.; Sprechmann, P.; and LeCun,
  Y.
\newblock 2016.
\newblock Disentangling factors of variation in deep representation using
  adversarial training.
\newblock In {\em NIPS},  5040--5048.

\bibitem[\protect\citeauthoryear{Mikolov \bgroup et al\mbox.\egroup
  }{2013}]{mikolov2013distributed}
Mikolov, T.; Sutskever, I.; Chen, K.; Corrado, G.~S.; and Dean, J.
\newblock 2013.
\newblock Distributed representations of words and phrases and their
  compositionality.
\newblock In {\em NIPS},  3111--3119.

\bibitem[\protect\citeauthoryear{Rao and Tetreault}{2018}]{rao2018dear}
Rao, S., and Tetreault, J.
\newblock 2018.
\newblock Dear sir or madam, may i introduce the {GYAFC} dataset: Corpus,
  benchmarks and metrics for formality style transfer.
\newblock In {\em NAACL}, volume~1,  129--140.

\bibitem[\protect\citeauthoryear{Rumelhart, Hinton, and
  Williams}{1985}]{rumelhart1985learning}
Rumelhart, D.~E.; Hinton, G.~E.; and Williams, R.~J.
\newblock 1985.
\newblock Learning internal representations by error propagation.
\newblock Technical report, California Univ San Diego La Jolla Inst for
  Cognitive Science.

\bibitem[\protect\citeauthoryear{Shen \bgroup et al\mbox.\egroup
  }{2017}]{shen2017style}
Shen, T.; Lei, T.; Barzilay, R.; and Jaakkola, T.
\newblock 2017.
\newblock Style transfer from non-parallel text by cross-alignment.
\newblock In {\em NIPS},  6833--6844.

\bibitem[\protect\citeauthoryear{Stent, Marge, and
  Singhai}{2005}]{stent2005evaluating}
Stent, A.; Marge, M.; and Singhai, M.
\newblock 2005.
\newblock Evaluating evaluation methods for generation in the presence of
  variation.
\newblock In {\em Int. Conf. Intelligent Text Processing and Computational
  Linguistics},  341--351.

\bibitem[\protect\citeauthoryear{Sutskever, Vinyals, and
  Le}{2014}]{sutskever2014sequence}
Sutskever, I.; Vinyals, O.; and Le, Q.~V.
\newblock 2014.
\newblock Sequence to sequence learning with neural networks.
\newblock In {\em NIPS},  3104--3112.

\bibitem[\protect\citeauthoryear{Tieleman and
  Hinton}{2012}]{tieleman2012lecture}
Tieleman, T., and Hinton, G.
\newblock 2012.
\newblock Lecture 6.5-rmsprop: Divide the gradient by a running average of its
  recent magnitude.
\newblock {\em COURSERA: Neural Networks for Machine Learning}.

\bibitem[\protect\citeauthoryear{Zhao \bgroup et al\mbox.\egroup
  }{2018}]{zhao2018adversarially}
Zhao, J.~J.; Kim, Y.; Zhang, K.; Rush, A.~M.; and LeCun, Y.
\newblock 2018.
\newblock Adversarially regularized autoencoders.
\newblock In {\em ICML},  5897--5906.

\end{thebibliography}
\bibliographystyle{aaai}

\pagebreak

\onecolumn

\appendix

\section{Supplemental Material}

\subsection{A. Bag-of-Words (BoW) Vocabulary Ablation Tests}

The tests in Table \ref{tab:bow-vocab-ablation} demonstrate the effect of the choice of vocabulary used for the auxiliary content losses.

\begin{table}[ht]
	\centering
	\begin{tabular}{| l || c | c | c | c |}
		\hline
		\tabc{2}{BoW Vocabulary}                         & \tabh{Transfer} & \tabh{Cosine}     & \tabh{Word}    & \tabh{Language} \\
		                                                 & \tabh{Strength} & \tabh{Similarity} & \tabh{Overlap} & \tabh{Fluency}  \\
		\hline
		\hline
		Full Corpus Vocabulary                           & 0.822           & 0.896             & 0.344          & -10.13          \\
		\hline
		Vocabulary without sentiment words               & 0.872           & 0.901             & 0.359          & -10.33          \\
		\hline
		Vocabulary without stopwords                     & 0.836           & 0.894             & 0.429          & -10.06          \\
		\hline
		Vocabulary without stopwords and sentiment words & \textbf{0.934}  & \textbf{0.904}    & \textbf{0.473} & \textbf{-9.84}  \\
		\hline
	\end{tabular}
	\caption{Ablation tests on the BoW vocabulary.}
	\label{tab:bow-vocab-ablation}
\end{table}

It is evident that using a BoW vocabulary that excludes sentiment words and stopwords performs better on every single quantitative metric.

\subsection{B. t-SNE plots of Ablation Tests}

\begin{figure}[ht]
	\includegraphics[width=\linewidth]{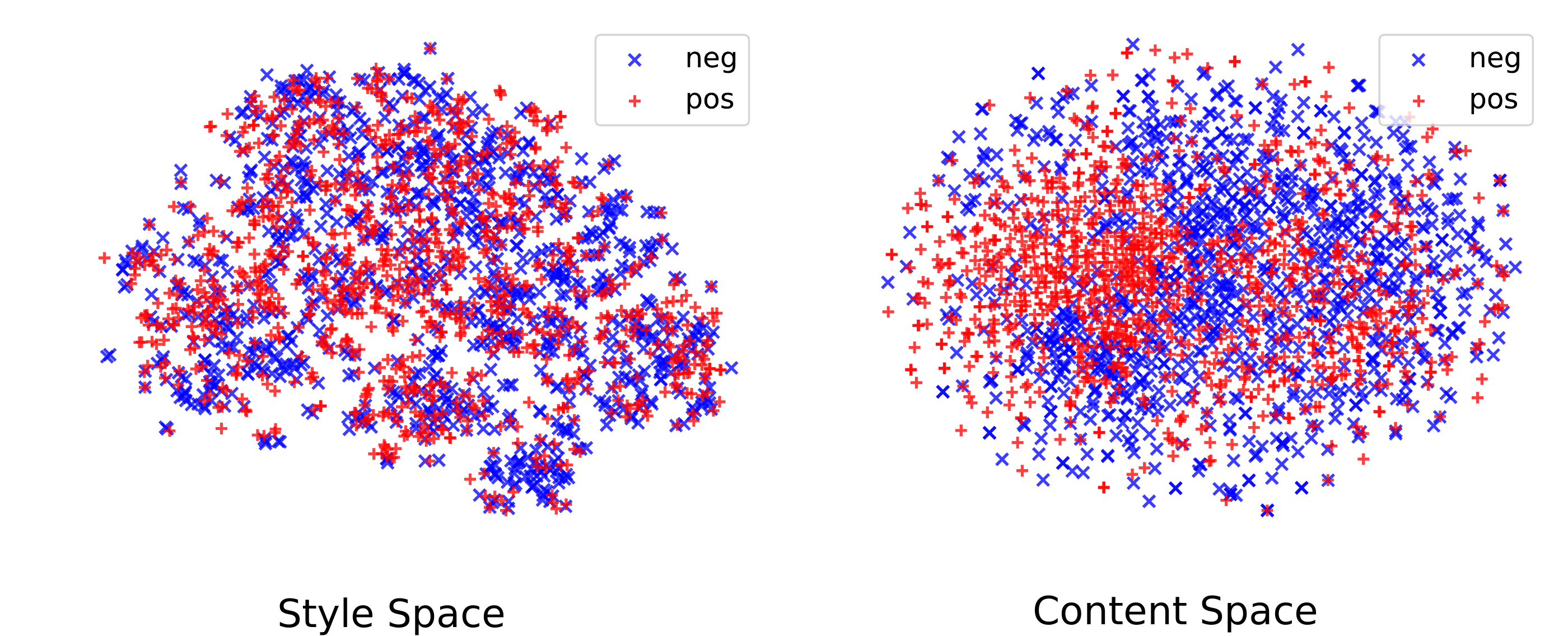}
	\caption{t-SNE Plot of VAE latent embeddings with only $\loss{AE}$.}
	\label{fig:only-rec}
\end{figure}

\begin{figure}[ht]
	\includegraphics[width=\linewidth]{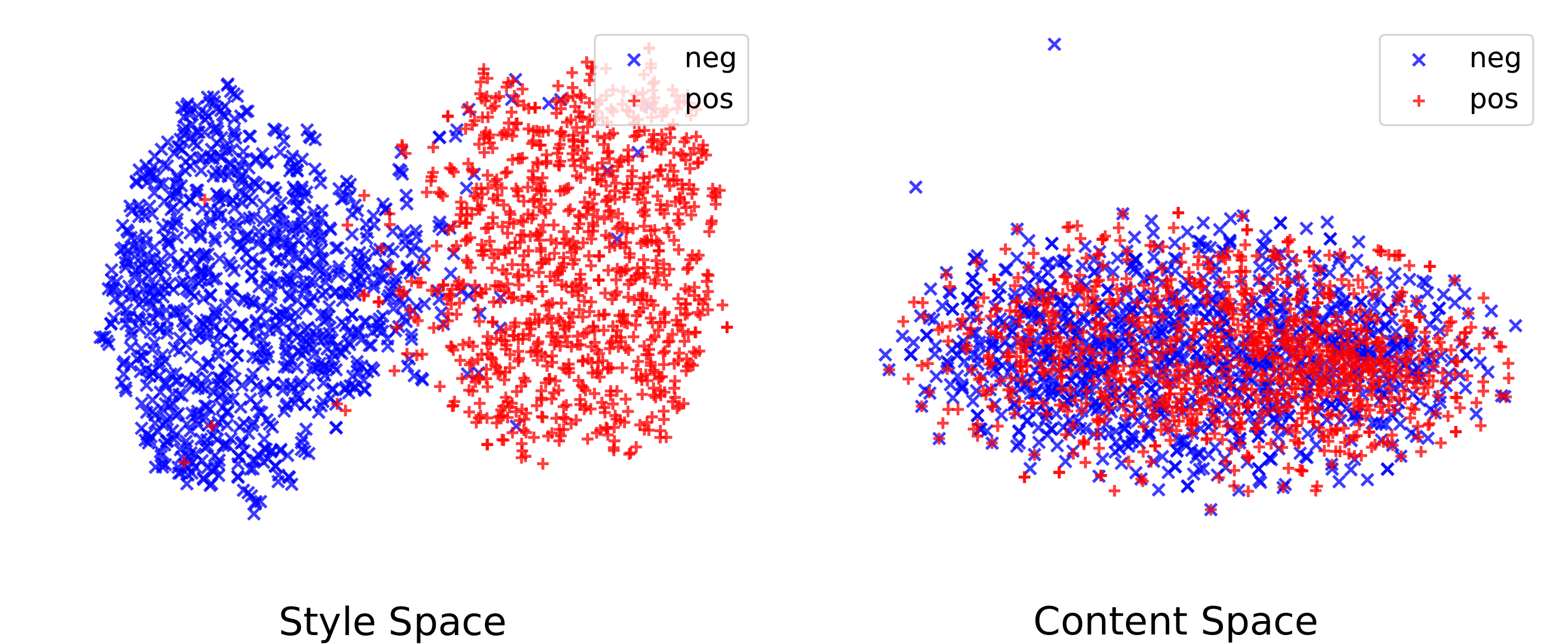}
	\caption{t-SNE Plot of VAE latent embeddings with $\loss{AE} + \loss{mul(s)}$.}
	\label{fig:rec-and-muls}
\end{figure}

\begin{figure}[ht]
	\includegraphics[width=\linewidth]{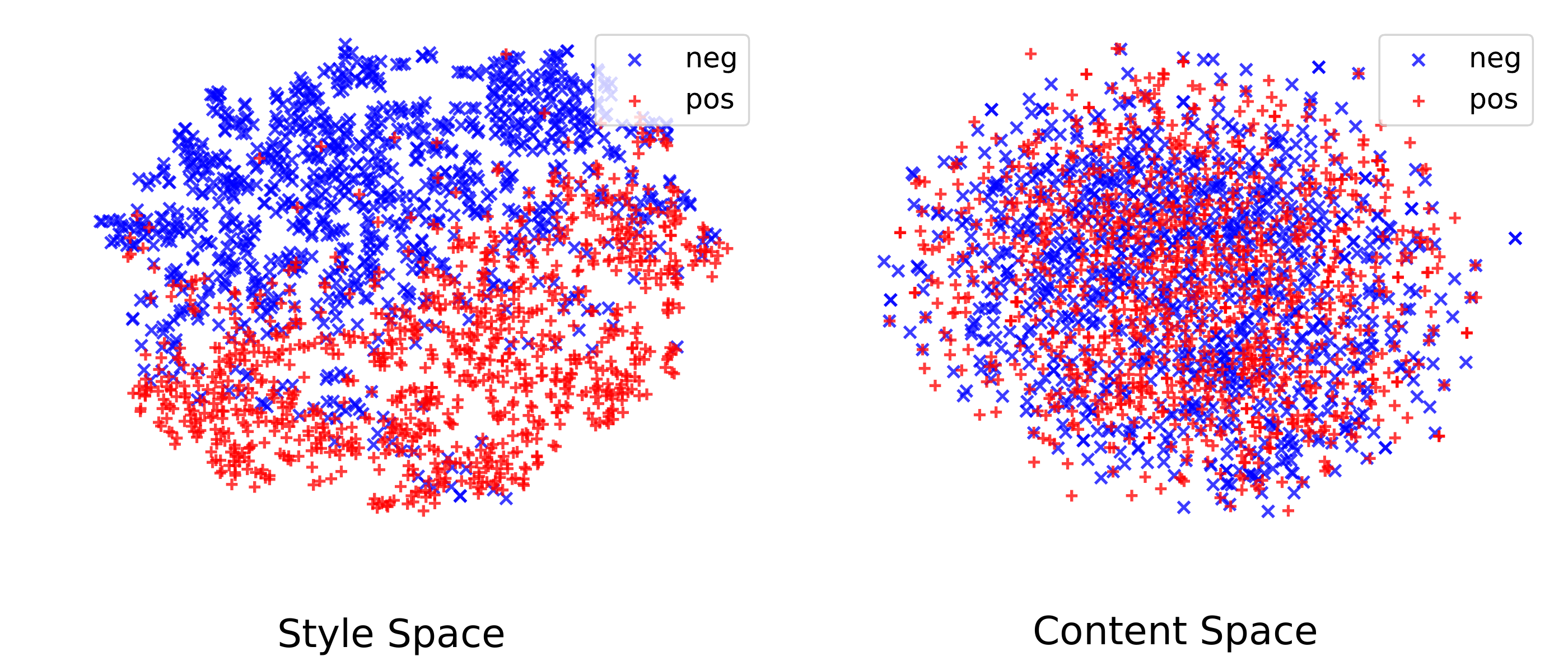}
	\caption{t-SNE Plot of VAE latent embeddings with $\loss{AE} + \loss{adv(s)}$.}
	\label{fig:rec-and-advs}
\end{figure}

\begin{figure}[ht]
	\includegraphics[width=\linewidth]{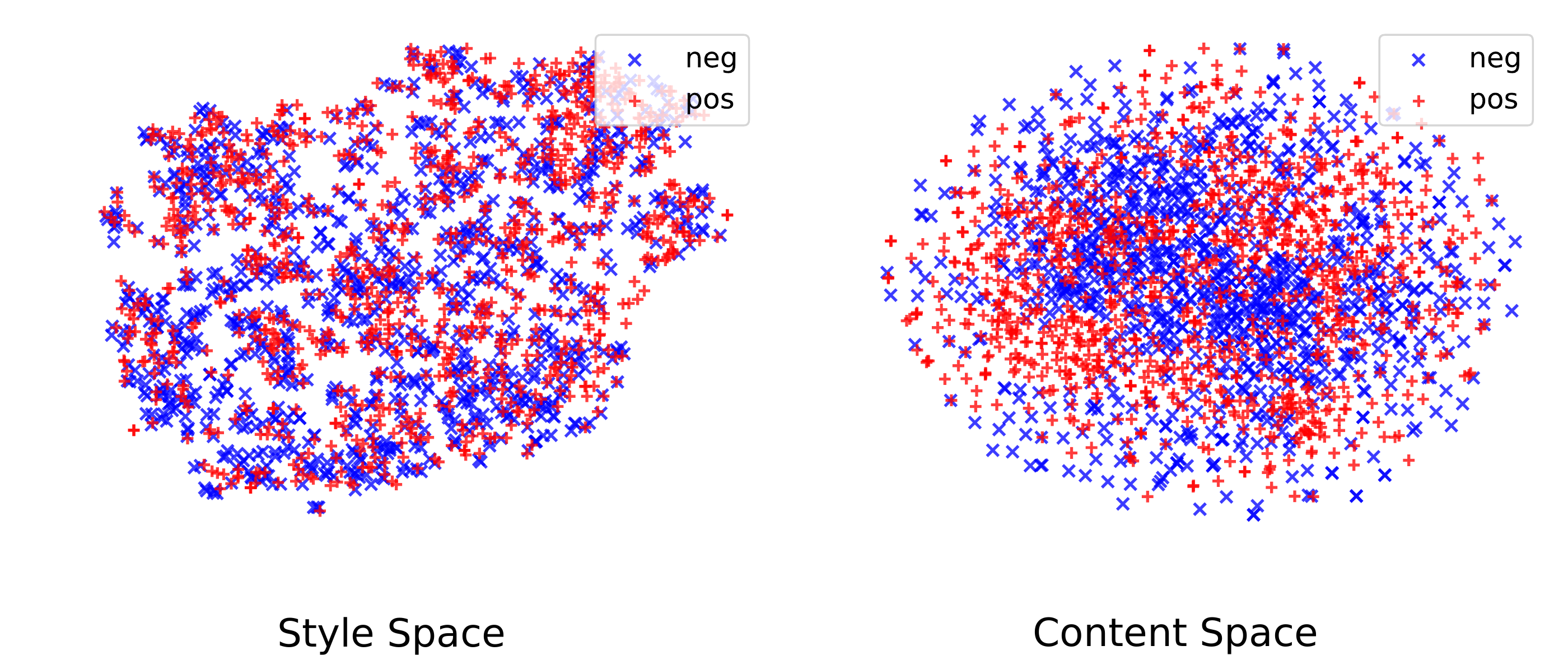}
	\caption{t-SNE Plot of VAE latent embeddings with $\loss{AE} + \loss{mul(c)}$.}
	\label{fig:rec-and-mulc}
\end{figure}

\begin{figure}[ht]
	\includegraphics[width=\linewidth]{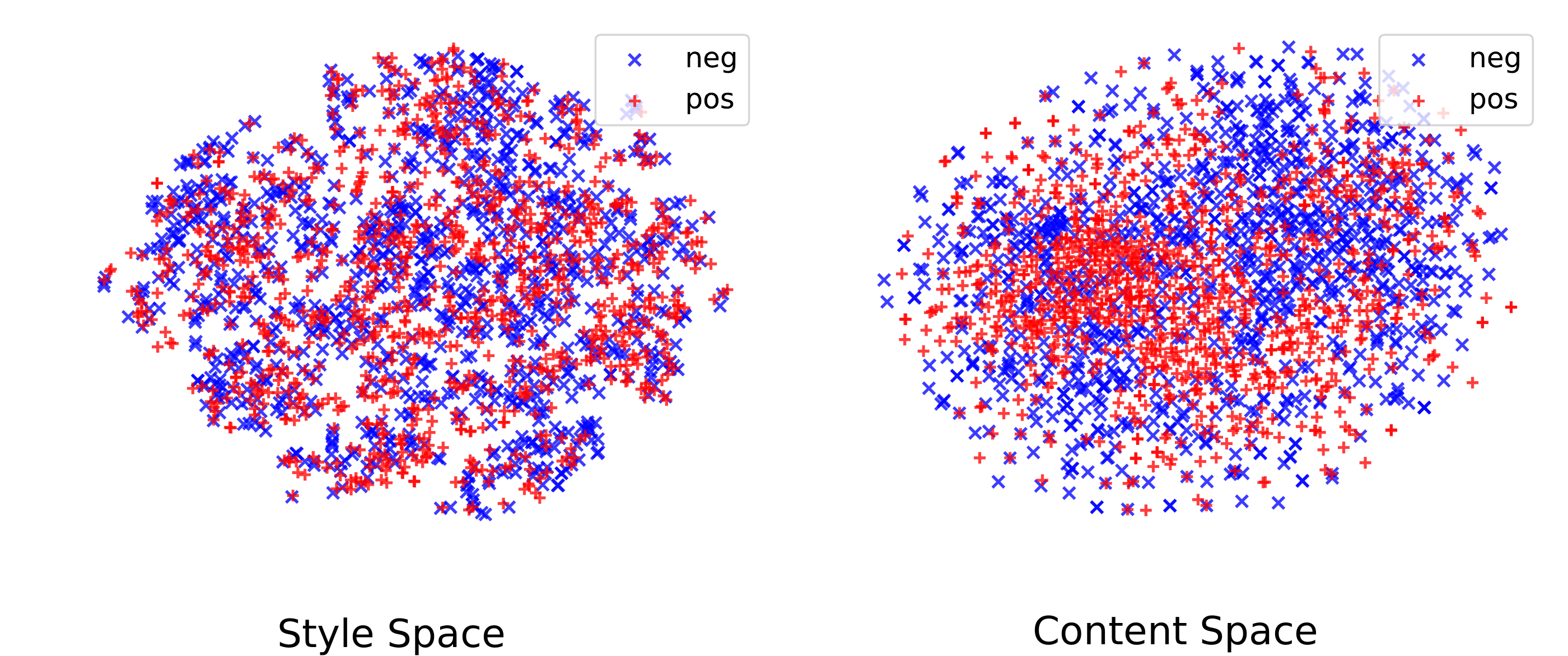}
	\caption{t-SNE Plot of VAE latent embeddings with $\loss{AE} + \loss{adv(c)}$.}
	\label{fig:rec-and-advc}
\end{figure}

Figure \ref{fig:only-rec} shows the t-SNE plots of the style and content embeddings, without any auxiliary losses.
Figures \ref{fig:rec-and-muls}, \ref{fig:rec-and-advs}, \ref{fig:rec-and-mulc} and \ref{fig:rec-and-advc} show the effect of adding each of the auxiliary losses independently.

\end{document}